\documentclass{INTERSPEECH2023}
\usepackage{multirow}

\usepackage{cite}

\usepackage{amssymb}
\usepackage{pifont}
\newcommand{\cmark}{\ding{51}}%
\newcommand{\xmark}{\ding{55}}%

\title{Phonetic and Prosody-aware Self-supervised Learning Approach for Non-native Fluency Scoring}

\name{Kaiqi Fu, Shaojun Gao, Shuju Shi, Xiaohai Tian, Wei Li, Zejun Ma}

\address{
  ByteDance}
\email{kaiq.fu@gmail.com, \ gaoshaojun123@163.com,\ \{shuju.shi,xiaohai.tian,liwei.speech,mazejun\}@bytedance.com}

\begin{document}

\maketitle
 
\begin{abstract}

Speech fluency/disfluency can be evaluated by analyzing a range of phonetic and prosodic features. Deep neural networks are commonly trained to map fluency-related features into the human scores. However, the effectiveness of deep learning-based models is constrained by the limited amount of labeled training samples. To address this, we introduce a self-supervised learning (SSL) approach that takes into account phonetic and prosody awareness for fluency scoring. Specifically, we first pre-train the model using a reconstruction loss function, by masking phones and their durations jointly on a large amount of unlabeled speech and text prompts. We then fine-tune the pre-trained model using human-annotated scoring data. Our experimental results, conducted on datasets such as Speechocean762 and our non-native datasets, show that our proposed method outperforms the baseline systems in terms of Pearson correlation coefficients (PCC). Moreover, we also conduct an ablation study to better understand the contribution of phonetic and prosody factors during the pre-training stage.

\end{abstract}
\noindent\textbf{Index Terms}: Computer Assisted Pronunciation Training (CAPT), Non-native Fluency Scoring, Phonetic and Prosody-aware, Self-suprevised Learning

\section{Introduction}
The ability to speak fluently is a significant aspect when evaluating a learner's language proficiency \cite{ex0}. It is characterized by the seamless and effortless production of speech with minimal pauses, hesitation, or corrections \cite{ex1,ex2,ex3,ex4}. L2 learners typically exhibit slower speech and more frequent unnecessary pauses compared to native speakers. Automatic scoring of fluency, serves as an essential module in computer-aided language learning (CALL) systems. It has been extensively studied in both ``read aloud" \cite{read1,nancy,gopt,3m,our,ssl1} and ``open response" \cite{open1,open2,open3,liuwei} scenarios. In the read aloud" scenario, L2 learners are required to read a provided prompt text, whereas the `open response" requires them to express their opinions freely based on a given question.

In this paper, we focus on ``read aloud" scenario, where forced-alignment model is first applied to a pair of non-native speech and prompt text to generate time stamps of speech segments, such as phonemes, words and etc. Fluency related features are then extracted and fed into subsequent fluency scorers. Although recent end-to-end neural network based fluency scorers have achieved satisfactory results~\cite{nancy,our,gopt,3m,ssl1,liuwei}, their performances heavily rely on the size of labeled scoring samples. In fact, the non-native data labeling process is costly and has scalability issues~\cite{lian}. Take the recently released public free dataset Speechocean762~\cite{speechocean} for example, only 5,000 sentences have been assigned with human fluency scores. The comparison in~\cite{nancydata} shows that the largest nonnative corpus only contains 90,841 utterances, but it is not publicly available. 

To overcome the challenge of limited labeled data, many researchers are using pre-training and fine-tuning paradigms to leverage large amounts of unlabeled data~\cite{ylf_pet,peng_mdd}. In the field of natural language processing (NLP), masked language modeling (MLM) has become a popular method for pre-training models such as BERT~\cite{bert}, RoBERTa~\cite{roberta}, and ERNIE~\cite{ernie}. MLM involves masking a subset of tokens in a sequence and training the model to predict these masked tokens, which enables the model to learn high-level contextual representations that can be beneficial for downstream tasks.

Recently, a new multi-stream transformer language model (MS-TLM) was proposed to jointly model phonetic content and prosody~\cite{glsm}, which demonstrated the effectiveness of prosody prediction. In this paper, we propose a self-supervised learning approach that incorporates phonetic and prosodic information to improve non-native fluency scoring. The pre-trained model is used to predict masked phones and durations, which enhances the model's ability to represent long-range phonetic and prosodic information. Specifically, we use an automatic speech recognition (ASR) system to generate phone-level raw sequential features, e.g. acoustic features, phone sequences, and duration, for pairs of non-native speech and prompt text. We then randomly mask 15\% of these phone-level features and train our fluency scorer to predict the masked  phone and duration. Finally, the pre-trained model is fine-tuned using limited human-annotated fluency scores. Experimental results show that the proposed approach can significantly improve fluency scoring in various configurations. An ablation study is also conducted to analyze the effect of different loss functions used in our pre-training stage on fluency scoring.

\begin{figure*}[t]
\centering

\includegraphics[width=1\textwidth]{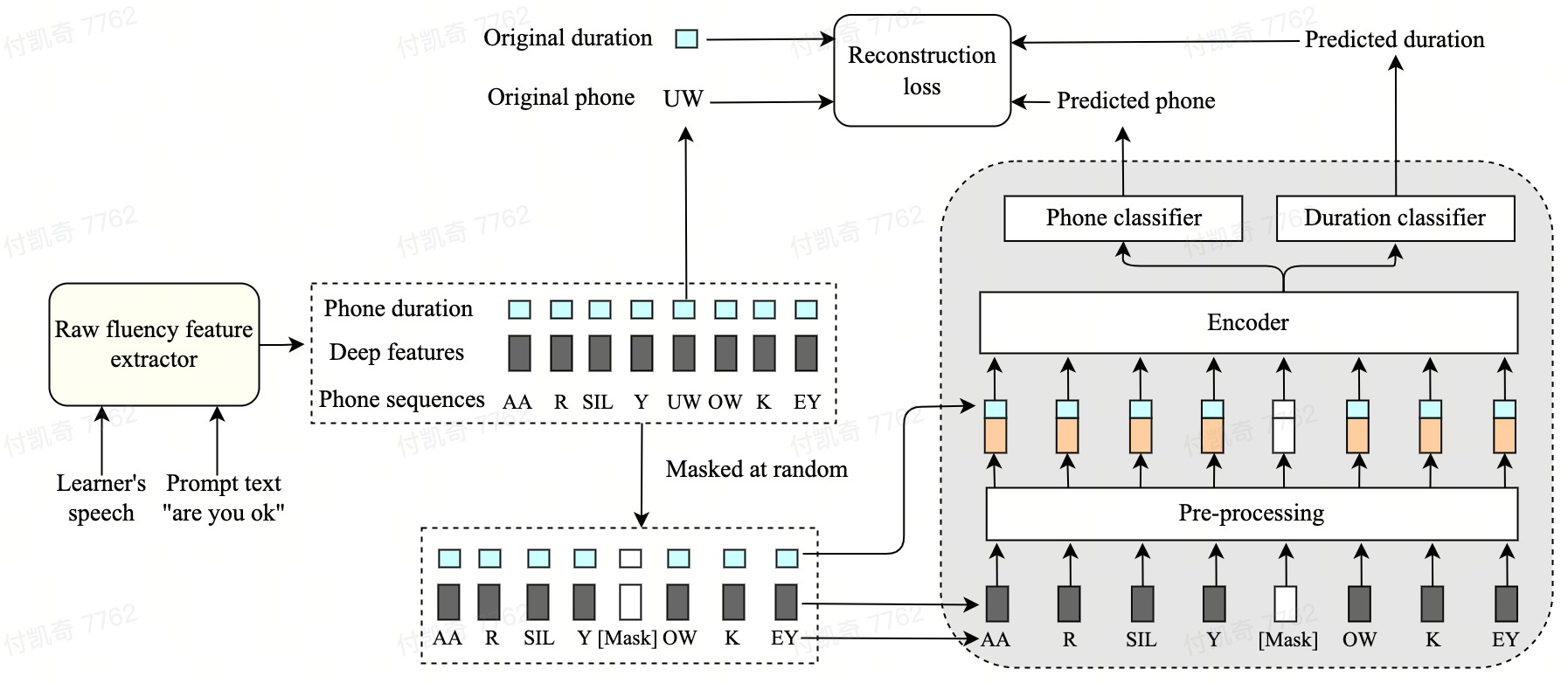}
\caption{An overview of the phonetic and prosody-aware pre-training process.}
\label{fig:method}
\end{figure*}

\section{Related work}

Over the past few decades, extensive research has been conducted on spoken fluency scoring. Traditionally, handcrafted features such as the statistics of speech break~\cite{read1}, speech rate~\cite{ read1, nancy, open1, open2, open3}, filled pause, and goodness of pronunciation (GOP)~\cite{nancy,gopt,3m} were collected based on phone boundaries and fed into various fluency scorers such as SVM~\cite{open1,open3}, and multiple linear~\cite{read1}. Recent works have employed sequence models to directly learn utterance-level fluency scores from phone-level raw features, including phonetic features (e.g., phone sequence~\cite{nancy,gopt,our,3m}), prosodic features (e.g., energy~\cite{3m}, pitch~\cite{nancy} and phone duration~\cite{our}. Bi-directional Long Short Term Memory (BLSTM)~\cite{nancy,our,liuwei,ssl1} and Transformer models~\cite{gopt,3m} have been used to capture the dynamic changes of phone-level pronunciation-related features for better modeling the evolution of local fluency over time.

More recently, self-supervised learning (SSL)-based speech models such as wav2vec2~\cite{wav2vec2} have been shown to be effective in learning meaningful representations from raw speech signals in various downstream tasks~\cite{sslall}. Inspired by this success, researchers used pre-trained SSL models like wav2vec2~\cite{wav2vec2}, HuBERT~\cite{hubert}, and WavLM~\cite{wavlm} to extract features directly and feed them into fluency scorers~\cite{ssl1,3m,liuwei}. Due to the promising performance, we consider the two SSL-based models~\cite{3m,liuwei} as strong baselines of this work.

\section{Method}
This section details our approach for fluency scoring. Initially, we outline our phonetic and prosody-aware pre-training technique that employs self-supervised learning to reconstruct the masked phone and duration in each pre-training sample. The process flow for pre-training is illustrated in Figure \ref{fig:method}. Subsequently, we elaborate on how we employ the pre-trained model for fluency scoring in downstream tasks.

\subsection{Phonetic and prosody-aware pre-training}

\subsubsection{Phone-level fluency feature extraction}
In \cite{our}, phone-level raw features were shown to be effective for assessing learners' speech fluency. Followed by the previous study, three segmental features (deep features, phone, and its duration) were extracted in this study. Initially, forced alignment is carried out to obtain time boundaries at the phone level, such as the beginning and end time of each phone and pause. The boundary information is then used to derive the phone sequence and its duration, which are denoted as $\mathbf{e} \in \textbf{R}^{\text{1} \times \text{N}}$ and $\mathbf{t} \in \textbf{R}^{\text{1} \times \text{N}}$, respectively. $\mathbf{t}$ refers to the number of frames within the phone. Following that, frame-level deep features (also known as bottleneck features) extracted from the acoustic model are averaged by the duration of each phone to obtain phone-level deep features, represented as $\mathbf{X} \in \textbf{R}^{\text{N} \times \text{D}}$, where $\text{D}$ and $\text{N}$ represent the acoustic feature dimension and the number of phones, respectively. Finally, the phone-level raw features are inputted into the model for pre-training and fine-tuning.

\subsubsection{Masking strategy}
During the pre-train stage, we utilize random masking to randomly replace 15\% of phone-level raw features in each sample with a special mask token. Among the selected positions, 90\% will be replaced with the special mask token and the remaining 10\% will be kept unchanged. The model takes three input features, which correspond to three different mask marker methods: 1) the selected phonemes are replaced with the mask token, 2) the selected phone duration is set to zero, and 3) the selected deep features are replaced with zero vectors. To provide the model with duration ground-truth, we set the duration label to a range of 1-100. If the phone duration exceeds 100 frames, we cap the duration label at 100.

\subsubsection{Multitask based reconstruction loss}


The pre-processing steps takes phone-level deep features $\mathbf{X}$ as input, which are initially transformed into a condensed feature space $\mathbf{X'}$ using a fully connected layer. Next, the phone sequence $\mathbf{e}$ is converted into phone embeddings $\mathbf{E}$. The sum of $\mathbf{E}$ and $\mathbf{X'}$ output is then concatenated with phone duration $\mathbf{t}$ and utilized as a sequence of input features for the SSL encoder, which generates the phone-level hidden representations $\mathbf{H}$. 

\begin{equation}
\label{eq:scoring2}
\mathbf{H} =  \mathcal{E}([ \mathbf{X'}+\mathbf{E}; \mathbf{t} ]),
\end{equation}
where $\mathcal{E}$ is presented the SSL encoder.



The phone-level hidden representations $\mathbf{H}$ are then passed through two classifiers for phoneme and duration prediction, respectively. The pre-training model is optimized jointly by utilizing a multitask approach that minimizes the cross-entropy loss between the predicted and ground truth phonemes and durations. The loss function of $i$-th masked token is described as follows:

\begin{equation}
\label{eq:loss}
\mathcal{L}_{i} = \mathcal{L}_{ce} ( {y}^p_i, \mathcal{P}_p(\mathbf{h}_i)) +\mathcal{L}_{ce} ({y}^d_i,  \mathcal{P}_d(\mathbf{h}_i)),
\end{equation}

where ${y}^p_i$ and ${y}^d_i$ are presented the ground truth phoneme and duration, respectively. $\mathbf{h}_i$ is the phone-level hidden representations in $i$-th masked position. The phoneme and duration classifiers are denoted as $\mathcal{P}_p(\cdot)$ and $\mathcal{P}_d(\cdot)$, respectively. It should be noted that the loss function is calculated exclusively based on the masked phonemes and duration. The total loss is calculated by summing the loss values of all the masked tokens across sentences.

\subsection{Fine-tuning for fluency scoring}

In this phase, our objective is to fine-tune the pre-trained model for fluency scoring. The fluency scoring model comprises an encoder and a scorer. Initially, we utilize the pre-trained weights to initialize the encoder of the scoring model. Subsequently, the scorer performs average pooling on a sequence of encoder outputs $\mathbf{H}$ (as illustrated in Eq.~(\ref{eq:scoring2})), resulting in an utterance-level fluency representation. This representation is then fed into a linear layer to generate machine score. Mean square error (MSE) calculated between predicted and human-annotated fluency scores are used as the objective for entire network fine-tuning.

\begin{table}[t!]
\vspace{1pt}
\centering
\caption{Data splitting for fluency scorer}
 \label{table:data}
\begin{tabular}{clcll}
\toprule
\multicolumn{1}{l}{}      &      & \textbf{Train}   & \textbf{Dev}   & \textbf{Test}    \\ \midrule
\textbf{Pre-train}                  & Unlabeled data & 203,206      & 2,000     & -     \\  \midrule
\multirow{2}{*}{\textbf{Fine-tune}} & ByteRead        & 10,000 & 2,000 & 2,000 \\
                          & Speechocean762  & 2,500  & -     & 2,500 \\ \bottomrule
\end{tabular}
\end{table}

\section{Experimental setup}
\label{sec:exp_setup}
\subsection{Speech corpora}
\label{ssec:dataset_setup}

The acoustic model was trained on a total of 5,000 hours of English speech data, including 960 hours of native speech from the LibriSpeech~\cite{librispeech} and 4,000 hours of non-native private recordings from Bytedance. Additionally, we collected approximately 436 hours (about 200,000 utterances) of reading speech by Chinese L2 adult learners and prompt text for MLM pre-training. To evaluate fluency scoring, we performed experiments on two additional datasets: ByteRead, an internal dataset of 14,000 English utterances collected from Bytedance's education product (described in detail in~\cite{our}), and Speechocean762, an open-source speech assessment corpus consisting of 5,000 utterances collected from 250 speakers \cite{speechocean}. The data statistics were detailed in Table \ref{table:data}.


\begin{table*}[t]
\vspace{1pt}
  \centering
  \caption{The PCC performance of different systems on ByteRead and Speechocean762 data sets.}
   \label{table:main_result}
\begin{tabular}{clcccc}
\toprule
     &    \multicolumn{1}{l}{Model}       & \#Param     & Pre-train  & Speechocean762 & ByteRead   \\ \midrule
     (a) &   BLSTM \cite{our}                 &     278K    &  -             & -                 & 0.817                        \\
   (b) &  3M-Transformer \cite{3m}                  &     -        &  -             & 0.828              &  -                          \\  
(c) &   SSL+IDX+BLSTM \cite{liuwei}                 &     -   &   -            &  0.795                 & 0.828                        \\ 
\midrule
  
   \multirow{2}{*}{(d)}  & \multirow{2}{*}{Transformer-pre}   & \multirow{2}{*}{795K}           & \xmark       & 0.784              & 0.783                 \\
                       &                                                       &                                 & \cmark       & 0.802              & 0.799                 \\ \midrule
    \multirow{2}{*}{(e)}  & \multirow{2}{*}{\textbf{BLSTM-pre}}                  &  \multirow{2}{*}{\textbf{871K}} & \xmark       & 0.797              & 0.804                 \\
                       &                                                        &                                 & \cmark       & \textbf{0.835}     & \textbf{0.833}        \\  \bottomrule

\end{tabular}
\end{table*}

\subsection{Feature extraction}
\label{ssec:feature setup}


Raw fluency features were extracted using the deep feedforward sequential memory network-hidden Markov models (DFSMN-HMM) acoustic model, as described in \cite{dfsmn}. The model architecture includes 2 convolutional layers, 24 FSMN layers, a bottleneck layer, and a feedforward layer. The input features were 39-dimensional Mel-frequency cepstral coefficients (MFCCs). The bottleneck layer extracts frame-level deep features with a dimensionality of 512. A HMM-based force-aligner is employed to obtain the phone sequence along with the corresponding start and end time boundaries for each phone.

\begin{table*}[t]
\vspace{3pt}
  \centering
  \caption{The PCC performance of BLSTM-based systems on different scales of the scoring training sets. ByteRead(1000) means 1,000 utterances with prompt text was randomly selected to fine-tune the pre-trained model in the ByteRead training set. Phn and dur loss represent phonetic and prosodic loss, respectively.}
   \label{table:ablation_result}

\begin{tabular}{lccccccc}
\toprule
\multicolumn{1}{l}{Model}    & Pre-train & Loss & ByteRead(1000) & ByteRead(2500) & ByteRead(5000) & ByteRead & Speechocean762 \\ \midrule
\multirow{4}{*}{\begin{tabular}[c]{@{}c@{}}BLSTM-pre\end{tabular}} & \xmark    & -           & 0.669   & 0.773          & 0.787         & 0.804   &  0.797    \\
                             & \cmark    & phn+dur   & \textbf{0.787}   & \textbf{0.807}         & \textbf{0.82}          & \textbf{0.833}  &  0.835   \\
                             & \cmark    & dur         & 0.78    & 0.8            & 0.818         & 0.826   &  \textbf{0.838}   \\
                             & \cmark    & phn         & 0.734   & 0.784          & 0.813         & 0.82    &  0.822   \\ \bottomrule
\end{tabular}
\end{table*}

\subsection{Setup of proposed and baseline systems}
\label{ssec:baselines}

\subsubsection{Proposed systems setup}
\label{sssec:proposed method}
Given the L2 learner‘s speech and prompt text, phone-level raw features (deep features,
phone sequence, and duration) can be first obtained in the fluency feature extraction module. The phone sequence is projected into a 32-dim phone embedding, while the 512-dim deep features are transformed into 32-dim features and added to the phone embedding to obtain the compact features. These compact features are then concatenated with a 1-dimensional duration feature, resulting in a 33-dimensional output of the pre-processing. This output serves as input to the pretrain model.

\begin{itemize}[leftmargin=*]
   
\item \textbf{Transformer-pre:} The proposed Transformer-based pretrain model. A trainable [CLS] token was appended to the processed feature sequence. And a trainable position embedding and the 33-dimensional processed features were summed together. The pre-trained encoder consists of two transformer layers, with the first layer removing the residual connection to increase the input feature dimension to 128. The multi-head attention block employs 4 heads. The output of the transformer encoder for the [CLS] token, with a dimensionality of 128, was used as the corresponding utterance-level representation for predicting the fluency score. 

\item \textbf{BLSTM-pre:} The proposed BLSTM-based pretrain model. The 64-dim phone-level contextual representations output of the BLSTM encoder will be fed into a mean pooling layer and a linear layer to get the final fluency score. According to our empirical study, an 8-layer BLSTM architecture lead to the best results. 

\end{itemize}
The adam optimization algorithm was employed for updating the pre-training and scoring models in all proposed systems. During pre-training, the batch size was set to 256 and the learning rate was 0.001. For fine-tuning, the batch size was set to 32 and the learning rate was 0.002.

\subsubsection{Baseline systems setup}
\begin{itemize}[leftmargin=*]

\item \textbf{BLSTM \cite{our}:} The baseline system without pretraining, where the scorer consists of pre-processing, a 2-layer BLSTM encoder, and a fully connected layer. The input feature and the pre-processing steps were the same as our proposed system as described in~\ref{sssec:proposed method}.

\item \textbf{3M-Transformer \cite{3m}:} The model input comprises multi-view phone-level features, which consist of prosodic features (duration, energy), SSL features (wav2vec2~\cite{wav2vec2}, HuBERT~\cite{hubert}, WavLM~\cite{wavlm}) , and GOP feature~\cite{hu2015improved}. These features are simply concatenated and subsequently fed into the 3-layer transformer-based scorer to get the fluency score. Multi-granularity pronunciation score labels are used to model the association between different scoring tasks.

\item \textbf{SSL-IDX-BLSTM \cite{liuwei}:} The system takes the frame-level SSL representations extracted from wav2vec2 Large as input. The k-means clustering algorithm is used to generate the clustered index, which is seen as pseudo phonetic information. A linear layer project the SSL feature into a compact feature, which is concatenated with the index embedding through an embedding layer and fed into 2-layer BLSTM to get the fluency score.

\end{itemize}

\section{Results and analyses}
\label{sec:res}
In our experiments, the system performance was evaluated using the Pearson correlation coefficient (PCC) between the machine-predicted scores and the human scores. A higher PCC value indicates a better system performance.


\subsection{Main results}
\label{ssec:self_config}

This subsection presents a comparison of the proposed method's performance on various encoder structures using the Speechocean762 and ByteRead datasets. Additionally, we assessed the proposed method's effectiveness by comparing its results with baselines. The results of the different systems are presented in Table~\ref{table:main_result}.

First, we evaluated the effectiveness of phonetic and prosody-aware pretrain models using both Transformer and BLSTM architectures, as shown in rows (d) and (e) of Table~\ref{table:main_result}. The results demonstrate that the proposed pretrain model methods consistently outperform their counterparts, which were trained from scratch with labeled data. This suggests that phonetic and prosody-aware pretraining can be beneficial for fluency scoring. Furthermore, we observed that the BLSTM pretrain model outperforms its Transformer counterpart. Hence, BLSTM pretrain model is used in the rest of the experiments.

Apart from the self implemented systems, we also conducted performance comparisons between the proposed BLSTM-pre system and the baseline systems presented in Table~\ref{table:main_result} (a), (b), and (c). 
We first compared the performance between proposed BLSTM-pre and the BLSTM system reported in \cite{our}. The results show that our BLSTM-pre outperformed the BLSTM baseline, with an improvement in PCC on the ByteRead database from 0.817 to 0.833. This confirms that the pretrain model is effective for fluency scoring.
We then conducted comparisons between our proposed BLSTM-pre system and two SSL feature-based approaches, namely 3M-Transformer and SSL-IDX-BLSTM. Our proposed BLSTM-pre system showed better performance than the 3M-Transformer baseline on the Speechocean762 database, resulting in an increase in PCC from 0.828 to 0.835. Similar results were observed in comparison with SSL-IDX-BLSTM, where our proposed BLSTM-pre system consistently achieved better performance on both Speechocean762 and ByteRead datasets. These findings suggest that our proposed system outperforms state-of-the-art systems for fluency scoring on both Speechocean762 and ByteRead datasets.

\subsection{Ablation studies}
\label{ssec:comparison_baseline}

In this section, we conducted a series of experiments to determine the relative importance of the phonetic and prosodic components in the proposed method for scoring fluency. We performed ablation studies by testing different loss function configurations and analyzing the performance of each component. The results are presented in Table 3.

Specifically, we first evaluated the pre-training model's performance by using only the phonetic aspect as the pre-training loss, which involved predicting the masked phone. Our findings revealed that this approach yielded better results than the no pre-training system. Moreover, we discovered that the prosodic aspect's contribution to the improvement was more substantial than that of the phonetic aspect. This could be attributed to the duration factor's significant role in assessing speech fluency. Finally, we combined both phonetics and prosody to optimize the pre-training model, resulting in a more significant improvement, highlighting the effectiveness of the proposed SSL method in fluency scoring.

\section{Conclusion}
This article introduces a self-supervised learning technique that is phonetic and prosody-aware for assessing the fluency of L2 learners' speech. The method involves masking the phone and duration of input features and then reconstructing them by utilizing a vast amount of unlabeled non-native data during the pre-training phase. To predict the fluency score, a small amount of scoring data was utilized to fine-tune the pre-trained model. Results based on the Speechocean762 datasets and our non-native dataset indicate that the proposed approach outperforms the baseline systems. Our future research aims to explore the benefits of our approach for scoring at various levels (such as phone, and word) and granularities (such as accuracy, and proficiency). Additionally, we plan to explore the impact of using the L1 dataset when pre-training.

\bibliographystyle{IEEEtran}
\bibliography{template}

\end{document}